\documentclass{bmvc2k}

\title{Temporal-controlled Frame Swap for Generating High-Fidelity Stereo Driving Data for Autonomy Analysis}

\addauthor{Yedi Luo}{}{1}
\addauthor{Xiangyu Bai}{}{1}
\addauthor{Le Jiang}{}{1}
\addauthor{Aniket Gupta}{}{2}
\addauthor{Eric Mortin}{}{3}
\addauthor{Hanumant Singh}{}{2}
\addauthor{Sarah Ostadabbas}{https://web.northeastern.edu/ostadabbas}{1}

\addinstitution{
 Augmented Cognition Lab,\\
 Electrical and Computer Engineering Department,
 Northeastern University,\\ Boston, MA, USA\\
}
\addinstitution{
 Field Robotics Lab,\\
 Electrical and Computer Engineering Department,
 Northeastern University,\\ Boston, MA, USA\\
}
 \addinstitution{
 DEVCOM Analysis Center (DAC)\\
 US Army
}

\runninghead{Luo et al.,}{Stereo Synthesis}

\usepackage{graphicx}
\usepackage[utf8]{inputenc}
\usepackage[T1]{fontenc}
\usepackage{bbding}
\usepackage{tikz}
\usepackage{comment}
\usepackage{amsmath,amssymb} 
\usepackage{color}
\usepackage{amssymb}
\usepackage{pifont}
\usepackage{booktabs}
\usepackage{multirow}
\usepackage{array}
\usepackage{tabu}
\usepackage{boldline}
\newcolumntype{Y}{>{\centering\arraybackslash}X}
\usepackage[accsupp]{axessibility}  

\newcommand{\figref}[1]{Fig.~\ref{fig:#1}}
\newcommand{\tabref}[1]{Table~\ref{tbl:#1}}

\newcommand{\overview}{
\begin{figure*}[t]
  \centering
  \includegraphics[width=\textwidth]{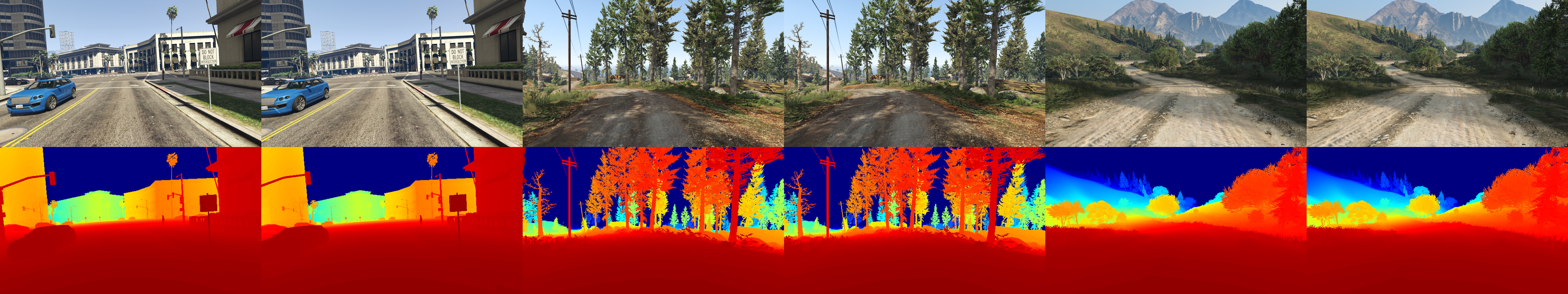}
  \caption{A few snapshots showcasing our GTAV-TeFS dataset, which includes synthetic stereo driving image pairs obtained from GTA V along with  various ground truth data such as depth maps, camera poses, GPS coordinates, temporal information, etc.}
    \vspace{-.2in}
  \label{fig:TeFS_sample}
\end{figure*}}

\newcommand{\process}{
\begin{figure*}[t]
  \centering
  \includegraphics[width=.9\textwidth]{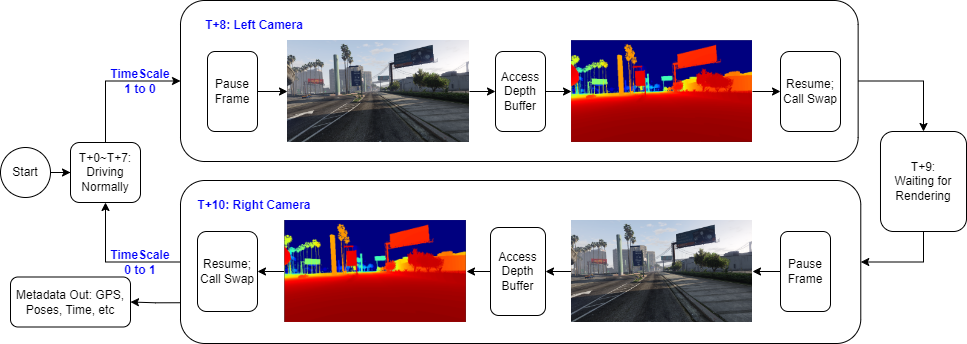}
  \caption{This flow chart outlines the Temporal-controlled Frame Swap (TeFS) algorithm for capturing synthetic stereo data from GTA V. It incorporates a single-thread architecture to precisely control the passage of time during the camera-swapping process. }
      \vspace{-.2in}
  \label{fig:TeFS_FC}
\end{figure*}}

\newcommand{\aaaview}{
\begin{figure*}[h]
  \centering
  \includegraphics[width=.9\textwidth]{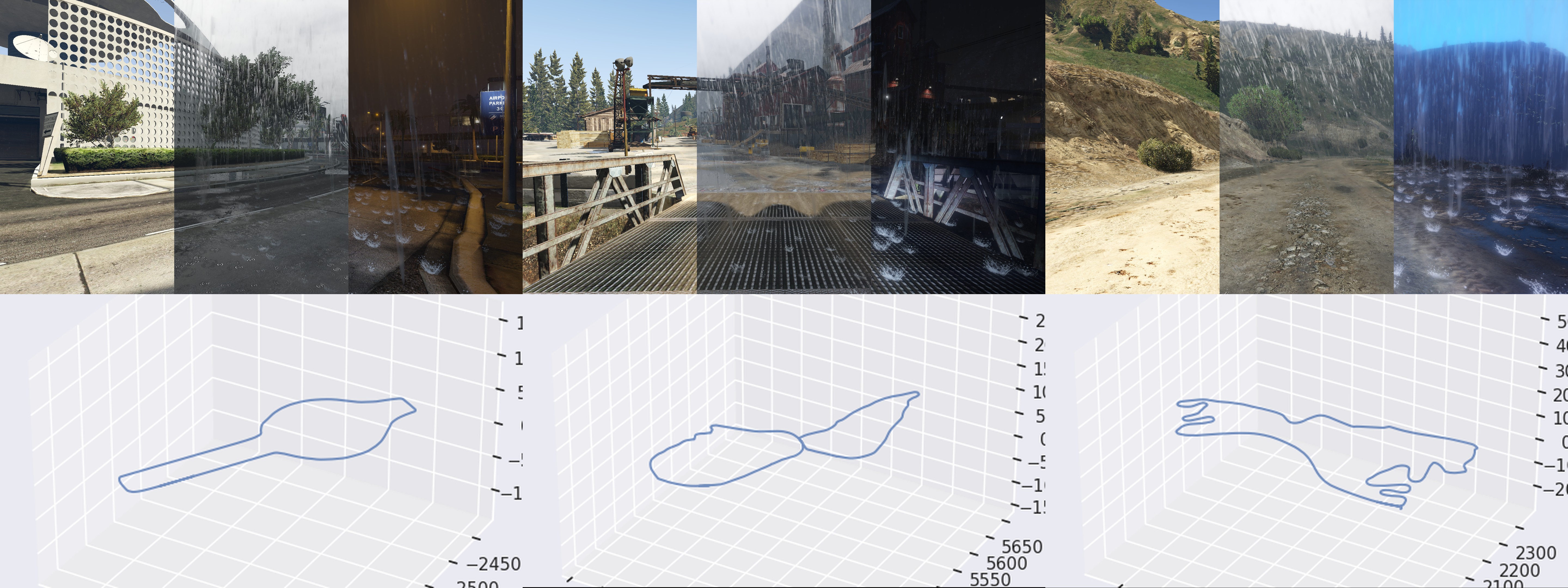}
  \caption{Comparison group samples: Three distinct road segments captured under diverse weather and lighting conditions, together with their corresponding ground truth trajectories.}
      \vspace{-.3in}
  \label{fig:cgroup}
\end{figure*}}

\newcommand{\cgcompare}{
\begin{figure*}[h]
  \centering
  \includegraphics[width=.9\textwidth]{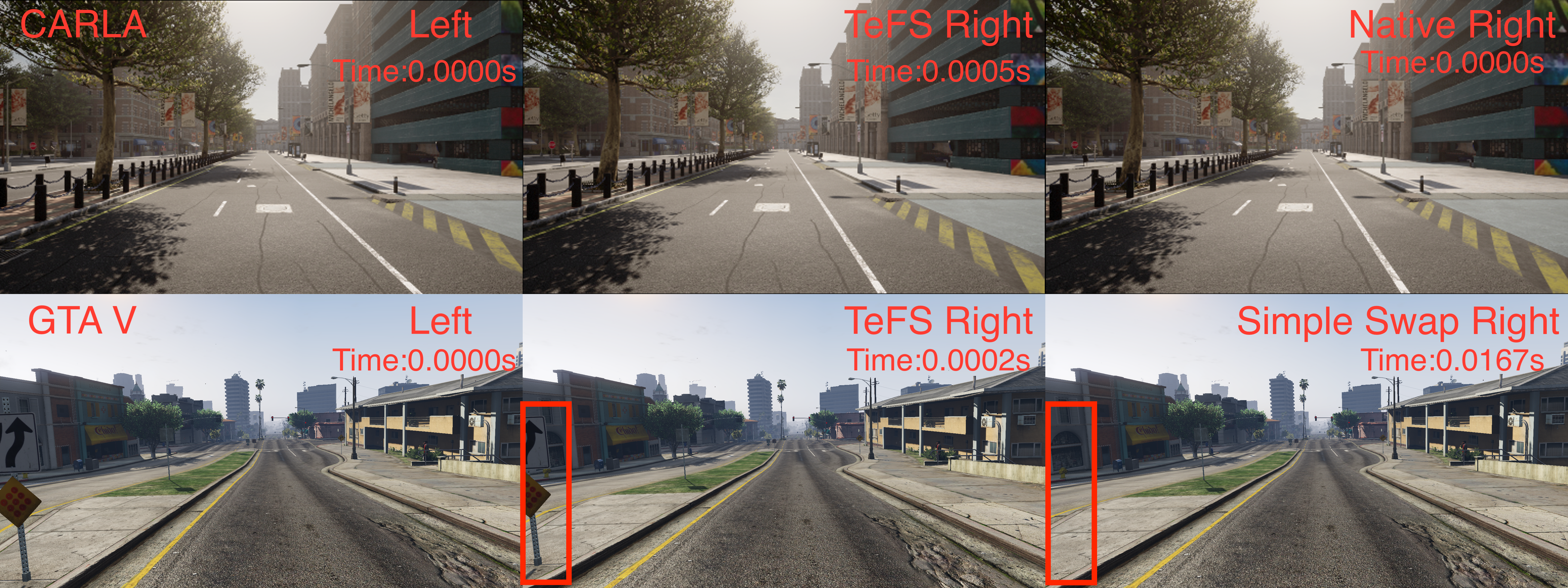}
  \caption{TeFS vs native stereo in CARLA and TeFS vs simple swap stereo in GTA V}
     \vspace{-.3in} 
  \label{fig:cgcompare}
\end{figure*}}
\newcommand{\carlaValidate}{
\begin{table*}[ht]
  \footnotesize
  \centering
  \caption{Stereo odometry evaluation results of datasets collected from CARLA \cite{CARLA} via TeFS method and the official stereo API, using ORB-SLAM3 \cite{ORBSLAM3}. In all experiments, we only aligned origin with ground truth. Results of our approach are marked in bold typeface, denoting performances that rival or sometimes surpass CARLA's native method.}
  \label{tbl:carlaValidate}
  \begin{tabular}{l l| l|c c c}
    \multirow{2}{*}{\textbf{Map}} & \multirow{2}{*}{\shortstack{\textbf{Trajectory} \\ \textbf{Distance (m)}}} & \multirow{2}{*}{\textbf{Method}} & 
    \multicolumn{3}{c}{\textbf{Evaluate Metrics}} \\
    \cline{4-6}
    &&& APE(m) $\downarrow$ & APE(\%) $\downarrow$ & RPE(m) $\downarrow$ \\ 
    \hline
    \multirow{2}{*}{Town 03} & \multirow{2}{*}{954.41} & Native & 1.67$\pm$0.59 & 0.17\%$\pm$0.06\% & 0.39$\pm$0.14 \\
    \cline{3-6}
    && TeFS  & \textbf{1.82$\pm$0.65} & \textbf{0.19\%$\pm$0.06\%} & \textbf{0.39$\pm$0.14} \\
    \cline{1-6}
    \multirow{2}{*}{Town 06} & \multirow{2}{*}{957.97} & Native & 4.22$\pm$3.17 & 0.44\%$\pm$0.33\% & 0.31$\pm$0.25 \\
    \cline{3-6}
    && TeFS & \textbf{2.30$\pm$1.67} & \textbf{0.24\%$\pm$0.17\%} & \textbf{0.31$\pm$0.25} \\
    \cline{1-6}
    \multirow{2}{*}{Town 10} & \multirow{2}{*}{950.05} & Native & 0.98$\pm$0.64 & 0.10\%$\pm$0.07\% & 0.27$\pm$0.22 \\
    \cline{3-6}
    && TeFS & \textbf{1.01$\pm$0.63} & \textbf{0.10\%$\pm$0.07\%} & \textbf{0.27$\pm$0.22} \\
    \cline{1-6}
  \end{tabular}
        \vspace{-.2in}
\end{table*}
}

\newcommand{\offroad}{
\begin{table*}[t]
\footnotesize
\caption{Stereo odometry evaluation results of various state-of-the-art feature-based vSLAM algorithms on the challenging-case comparison groups of our GTAV-TeFS dataset. In all experiments, we only aligned origin with ground truth. }
\vspace{-.5cm}
\begin{center}
\resizebox{\textwidth}{!}{
\begin{tabular}{cc|c|c|cccc}
\multirow{2}{*}{\textbf{Scene}} & \multirow{2}{*}{\textbf{Length(m)}} & \multirow{2}{*}{\textbf{Weather}} & \multirow{2}{*}{\textbf{Model}}  & \multicolumn{4}{c}{\textbf{Metrics}} \\
\cline{5-8}
&&&& APE(m)$\downarrow$   & APE(\%)$\downarrow$   & RPE(m)$\downarrow$ & Loop Detected \\
\hline
 \multirow{9}{*}{City 04} &  \multirow{9}{*}{760.0}  & \multirow{3}{*}{Extra Sunny}   &ORBSLAM3&3.20$\pm$0.85&0.42$\pm$0.11&2.78$\pm$0.16&\checkmark\\ \cline{4-8}
&&& OV$^2$SLAM&\textbf{2.53$\pm$1.21} & \textbf{0.33$\pm$0.16} & \textbf{0.06$\pm$0.09}&\checkmark\\\cline{4-8}
&&& VINS Fusion &2.95$\pm$1.36&0.39$\pm$0.18&0.09$\pm$0.26&\checkmark\\\clineB{3-8}{2.5}
&& \multirow{3}{*}{Cloudy with Rain}   &ORBSLAM3&\textbf{2.13$\pm$1.19} & \textbf{0.28$\pm$0.16} & \textbf{0.03$\pm$0.04} & \checkmark\\\cline{4-8}
&&& OV$^2$SLAM&21.01$\pm$9.33&2.77$\pm$1.23&0.29$\pm$2.28&\checkmark\\\cline{4-8}
&&& VINS Fusion &3.63$\pm$2.00&0.48$\pm$0.26&0.21$\pm$0.48&\checkmark\\\clineB{3-8}{2.5}
&& \multirow{3}{*}{Night Thunderstorm}   &ORBSLAM3&\multicolumn{4}{c}{Lost track of the map}\\\cline{4-8}
&&& OV$^2$SLAM & \textbf{54.19$\pm$48.40} & \textbf{7.11$\pm$6.35} & \textbf{0.59$\pm$4.15} \\\cline{4-8}
&&& VINS Fusion &\multicolumn{4}{c}{Map range unreachable}\\\hline\hline

\multirow{9}{*}{Offroad 01} &  \multirow{9}{*}{1206.0} & \multirow{3}{*}{Extra Sunny}   &ORBSLAM3 & \textbf{3.16$\pm$1.65} & \textbf{0.26$\pm$0.14} & \textbf{0.11$\pm$0.12} &\checkmark\\\cline{4-8}
&&& OV$^2$SLAM&4.81$\pm$2.21&0.39$\pm$0.18&0.21$\pm$0.19\\\cline{4-8}
&&& VINS Fusion &4.27$\pm$1.73&0.35$\pm$0.14&0.16$\pm$0.25&\checkmark\\\clineB{3-8}{2.5}
&& \multirow{3}{*}{Cloudy with Rain}   & ORBSLAM3 & \textbf{4.19$\pm$2.39} & \textbf{0.35$\pm$0.20} & \textbf{0.07$\pm$0.40} & \checkmark\\\cline{4-8}
&&& OV$^2$SLAM&6.60$\pm$4.19&0.54$\pm$0.34&0.20$\pm$0.25\\\cline{4-8}
&&& VINS Fusion &49.75$\pm$32.98&4.12$\pm$2.73&0.38$\pm$4.73&\checkmark\\\clineB{3-8}{2.5}
&& \multirow{3}{*}{Night Thunderstorm}   &ORBSLAM3&\multicolumn{4}{c}{Lost track of the map}\\\cline{4-8}
&&& OV$^2$SLAM&\textbf{91.90$\pm$45.50} & \textbf{7.57$\pm$3.79} & \textbf{0.29$\pm$2.31}\\\cline{4-8}
&&& VINS Fusion &\multicolumn{4}{c}{Map range unreachable}\\\hline\hline

 \multirow{9}{*}{Offroad 02}&  \multirow{9}{*}{3230.0} & \multirow{3}{*}{Extra Sunny}   & ORBSLAM3 & \textbf{8.19$\pm$3.95} & \textbf{0.25$\pm$0.12} & \textbf{0.39$\pm$0.26} &\checkmark\\\cline{4-8}
&&& OV$^2$SLAM&15.47$\pm$12.69&0.48$\pm$0.39&0.36$\pm$2.07&\checkmark\\\cline{4-8}
&&& VINS Fusion &15.44$\pm$8.99&0.48$\pm$0.39&0.16$\pm$0.21&\checkmark\\\clineB{3-8}{2.5}
&& \multirow{3}{*}{Cloudy with Rain}   & ORBSLAM3 & \textbf{9.45$\pm$2.77} & \textbf{0.29$\pm$0.09} & \textbf{0.16$\pm$0.14}\\\cline{4-8}
&&& OV$^2$SLAM&85.62$\pm$65.35&2.65$\pm$2.03&0.14$\pm$0.21\\ \cline{4-8}
&&& VINS Fusion &72.47$\pm$61.96&2.26$\pm$1.93&0.22$\pm$0.70\\\clineB{3-8}{2.5}
&& \multirow{3}{*}{Night Thunderstorm}   &ORBSLAM3&\multicolumn{4}{c}{Lost track of the map}\\\cline{4-8}
&&& OV$^2$SLAM & \textbf{160.33$\pm$82.39} & \textbf{4.97$\pm$2.49} & \textbf{0.34$\pm$3.54}\\\cline{4-8}
&&& VINS Fusion &\multicolumn{4}{c}{Map range unreachable}\\
\hline

\end{tabular}
}
\vspace{-.7cm}
\label{tbl:compare}
\end{center}
\end{table*}
}

\newcommand{\droid}{
\begin{table*}[t]
\footnotesize
\caption{Evaluation results of DROID-SLAM's \cite{teed2021droid}stereo and RGB-D odometry on the challenging-case comparison groups of our GTAV-TeFS dataset. The table includes both corrected APE/RPE scores with ground truth and raw APE results. The inclusion of raw APE results emphasizes DROID-SLAM's stereo scaling issue, while providing a comprehensive evaluation of its overall performance.}
\vspace{-.5cm}
\begin{center}
\resizebox{\textwidth}{!}{
\begin{tabular}{c|c|c|ccc|c}
\multirow{2}{*}{\textbf{Scene}} & \multirow{2}{*}{\textbf{Weather}} & \multirow{2}{*}{\textbf{Input Mode}}  & \multicolumn{3}{c}{\textbf{Metrics(Corrected)}} & \textbf{Metrics(Raw)} \\
\cline{4-7}
&&& APE(m)$\downarrow$  & APE(\%)$\downarrow$   & RPE(m)$\downarrow$ & APE(m)$\downarrow$ \\
\hline
\multirow{6}{*}{City 04}  & \multirow{2}{*}{Extra Sunny}  
&Stereo&1.27$\pm$0.58 & 0.23$\pm$0.12 & 0.63$\pm$0.14&73.69$\pm$33.71\\\cline{3-7}
&&RGBD& \textbf{0.86$\pm$0.45} & \textbf{0.11$\pm$0.06} & \textbf{0.63$\pm$0.10} &  \textbf{15.77$\pm$6.79}\\\clineB{2-7}{2.5}

& \multirow{2}{*}{Cloudy with Rain} 
&Stereo&\textbf{1.73$\pm$0.95} & \textbf{0.23$\pm$0.12} & 0.63$\pm$0.14 & 73.75$\pm$33.65\\\cline{3-7}
&& RGBD & 2.99$\pm$1.19 & 0.39$\pm$0.16 & \textbf{0.62$\pm$0.10} &\textbf{13.16$\pm$5.48}\\ \clineB{2-7}{2.5}

& \multirow{2}{*}{Night Thunderstorm} 
&Stereo&6.42$\pm$2.09 & 0.84$\pm$0.27 & 0.63$\pm$0.20  & 74.61$\pm$33.79 \\ \cline{3-7}
&& RGBD& \textbf{4.91$\pm$2.39} & \textbf{0.64$\pm$0.31} & \textbf{0.62$\pm$0.15} & \textbf{10.29$\pm$4.77}\\ 
\hline\hline

\multirow{6}{*}{Offroad 01}  & \multirow{2}{*}{Extra Sunny}  
&Stereo& 3.61$\pm$1.15 & 0.30$\pm$0.10 & 0.58$\pm$0.19 & 98.68$\pm$49.18\\\cline{3-7}
&&RGBD&\textbf{1.35$\pm$0.54} & \textbf{0.11$\pm$0.04} & \textbf{0.58$\pm$0.18} & \textbf{19.72$\pm$9.85}\\\clineB{2-7}{2.5}

& \multirow{2}{*}{Cloudy with Rain} 
&Stereo&6.23$\pm$3.66 & 0.51$\pm$0.30 & 0.59$\pm$0.18 & 99.44$\pm$49.29 \\\cline{3-7}
&& RGBD&\textbf{2.64$\pm$1.13} & \textbf{0.22$\pm$0.09} & \textbf{0.58$\pm$0.18} & \textbf{18.59$\pm$9.23}\\\clineB{2-7}{2.5}

& \multirow{2}{*}{Night Thunderstorm} 
&Stereo&{18.24$\pm$6.11} & {1.50$\pm$0.50} & {0.62$\pm$0.31} & 100.32$\pm$50.09 \\\cline{3-7}
&& RGBD&\textbf{6.68$\pm$4.42} & \textbf{0.55$\pm$0.35} & \textbf{0.62$\pm$0.19} & \textbf{21.48$\pm$12.89}\\ 
\hline

\end{tabular}
}
\vspace{-.7cm}
\label{tbl:droid}
\end{center}
\end{table*}
}

\begin{document}
\maketitle

\begin{abstract}
This paper presents a novel approach, TeFS (Temporal-controlled Frame Swap), to generate synthetic stereo driving data for visual simultaneous localization and mapping (vSLAM) tasks. TeFS is designed to overcome the lack of native stereo vision support in commercial driving simulators, and we demonstrate its effectiveness using Grand Theft Auto V (GTA V), a high-budget open-world video game engine. We introduce GTAV-TeFS, the first large-scale GTA V stereo-driving dataset, containing over 88,000 high-resolution stereo RGB image pairs, along with temporal information, GPS coordinates, camera poses, and full-resolution dense depth maps. GTAV-TeFS offers several advantages over other synthetic stereo datasets and enables the evaluation and enhancement of state-of-the-art stereo vSLAM models under GTA V's environment. We validate the quality of the stereo data collected using TeFS by conducting a comparative analysis with the conventional dual-viewport data using an open-source simulator. We also benchmark various vSLAM models using the challenging-case comparison groups included in GTAV-TeFS, revealing the distinct advantages and limitations inherent to each model. The goal of our work is to bring more high-fidelity stereo data from commercial-grade game simulators into the research domain and push the boundary of vSLAM models
\footnote{The simulated datasets and our code is available at \href{https://github.com/ostadabbas/Temporal-controlled-Frame-Swap-GTAV-TeFS-}{https://github.com/ostadabbas/Temporal-controlled-Frame-Swap-GTAV-TeFS-}. Video at: \href{https://www.youtube.com/watch?v=mIfOBIx3HwA&ab_channel=SarahOstadabbas}{https://www.youtube.com/watch?v=mIfOBIx3HwA\&ab\_channel=SarahOstadabbas}.}.
\end{abstract}

\section{Introduction}
Stereo vision in motion is crucial for autonomous-vehicle navigation research, enabling critical visual simultaneous localization and mapping (vSLAM)  tasks such as stereo odometry and up-to-scaled map creation. To enhance the robustness and performance of algorithms solving these tasks, large amounts of high-quality stereo driving data are required. Compared with expensive and hard-to-obtain real-world data, synthetic data has been increasingly used as a cost-effective and scalable solution. Popular open-source simulators, such as CARLA \cite{CARLA}, SYNTHIA \cite{synthia}, and AirSim \cite{airsim}, provide comprehensive application programming interfaces (APIs) for stereo data collection, significantly mitigating the scarcity of stereo driving data. However, due to their nonprofit nature, their environmental complexity and texture quality are not on par with today's commercial video games. A prime example is Grand Theft Auto V (GTA V) \cite{gtav2013}, a 265-million-dollar-budget \cite{gtav265} open-world game with superior realism and remarkable diversity that surpasses the capabilities of most open-source simulators. 

\overview

As a result, many research teams have constructed high-quality datasets from GTA V \cite{hurl2019presil,stewart2019ursa,DeepMVS,bai2023saves}, but, to date, no one has been able to provide a viable large-scale synthetic stereo-driving dataset collected from GTA V.
As a commercial video game, GTA V has shortcomings in ease of data collection. Unlike open-source platforms with built-in dual-viewport stereo collection API, closed-source video games like GTA V not only have no native stereo vision support but also pose a strict single viewport limitation to prohibit rendering different camera views simultaneously. 

To address this limitation and enable dynamic stereo data collection in all video games with similar restrictions, we introduce a novel method called Temporal-controlled Frame Swap (TeFS). To concretely illustrate the implementation details and effectiveness of our TeFS method, we use GTA V as our demonstration platform, and create the first large-scale GTA V-based stereo-driving dataset, GTAV-TeFS. Our dataset includes over 88,000 1920$\times$1080 high-resolution stereo RGB image pairs along with temporal information, GPS coordinates, camera poses, and full-resolution dense depth maps. Compared to the existing GTA V-based datasets, our dataset enables the evaluation and enhancement of stereo vSLAM tasks within the GTA V environment. Compared to other synthetic stereo datasets, GTAV-TeFS also offers several advantages, including closer-to-realism environmental complexity, scenarios in both city and off-road environments, and comparison groups that feature controlled road segments under various challenging cases. A few snapshots of our GTAV-TeFS dataset is shown in \figref{TeFS_sample}.

Considering TeFS collection process is different from the conventional dual-viewport stereo, we also prove that stereo data collected using TeFS is of similar quality to dual-viewport data.  Ideally, we would require a directly comparable dual-viewport stereo dataset from GTA V. However, no such dataset is currently available. Fortunately, TeFS is universally applicable, so we implemented TeFS in the open-source simulator CARLA \cite{CARLA}  to compare TeFS stereo data with native stereo data. Finally, we conduct an extensive evaluation of various vSLAM models, including both feature-based and learning-based approaches, utilizing the challenging-case comparison groups provided by our 'GTAV-TeFS' dataset. Our objective was to push the boundaries of state-of-the-art vSLAM models, effectively showcasing their distinct strengths and limitations. To summarize, the main contributions of our work are as follows:




\begin{itemize}
\item Introducing a novel method, Temporal-controlled Frame Swap (TeFS), to collect stereo-driving data from commercial video games with single viewport limitations.
\item Creating the first large-scale stereo driving dataset, GTAV-TeFS, based on the high-fidelity video game, GTA V, which includes stereo RGB image pairs, temporal information, GPS coordinates, camera poses, and full-resolution dense depth maps, facilitating up-to-scale stereo odometry tasks in a realistic commercial game environment.
\item Validating the quality of stereo data collected using TeFS by demonstrating comparable results in stereo odometry evaluations to that of traditional dual-viewport data.
\item Evaluating various vSLAM models using the challenging-case comparison groups offered by our GTAV-TeFS dataset, revealing the distinct advantages and limitations inherent to each model
\end{itemize}

\section{Related Work}

Numerous synthetic datasets have been created from the commercial game GTA V due to its high-quality graphics and complex environment. Authors in "Play For Data"  offered 25,000 video frames for training semantic segmentation systems \cite{Richter2016PlayingForData}, while in "Play For Benchmark", they provided over 250,000 high-resolution video frames annotated with ground-truth data for benchmarking multi-level vision tasks, such as tracking and perception. \cite{Richter_2017playforbenchmark}. Another example is the PreSIL dataset \cite{hurl2019presil}, which includes 50,000 MonoRGB image frames and corresponding LiDAR point cloud data, has also proven valuable in improving the accuracy of 3D Object Detection Benchmarks. However, these GTA V datasets have a shared limitation: the lack of stereo image pairs, which impedes their use in vSLAM tasks. With only the monocular data, the scaling issue of estimated trajectories is almost inevitable. Additionally, multi-view GTA datasets like GTA SfM \cite{sfm} and MVS-Synth \cite{DeepMVS} also suffer from their own shared limitations from a different perspective. While both provide multiple-angle RGB images of static scenes along with ground truth depth maps for 3D scene reconstruction, neither offers dynamic stereo image pairs in motion, rendering them unsuitable for stereo odometry tasks as well.

Apart from commercial games, open-source simulators are widely used for computer vision and autonomous driving research. One notable example is Microsoft's AirSim \cite{airsim}, a high-fidelity simulator that has been used to create the TartanAir dataset \cite{tartanair2020iros}, offering rich data modalities such as stereo RGB images, depth maps, semantic segmentation masks, and camera poses across diverse environments. Similarly, CARLA \cite{CARLA} simulator provides realistic urban environments with numerous ground truth labels and the ability to configure stereo camera setups for autonomous driving research. One significant advantage of these simulators is their comprehensive data collection API, as they were specifically designed for research purposes. 
Nonetheless, even though their data collection tools are more accessible, their geometry complexity and scene diversity do not match the level of realism provided by commercial-grade video games due to their significant budget gaps.

In conclusion, due to the limitations present in commercial games like GTA V and the reduced environmental quality of open-source simulators, there is a definite need for a universal approach that overcomes the single viewport restriction and allows for the acquisition of synthetic stereo driving data in commercial video games and driving simulators. 

\section{Synthetic Stereo Driving Data} 
This section provides a comprehensive overview of the implementation of our TeFS method in GTA V. We describe the interaction with the game environment, the stereo camera setup, the TeFS workflow, 
the process of obtaining depth maps, and the limitations of our approach.

\subsection{Modding and Camera Setup in GTA V} 
Our ability to interact with GTA V largely relies on the highly developed modding community, with the most important plugin being ScriptHook V \cite{ScriptHookV}. It enables us to directly interact with GTA V through API libraries such as ScriptHookVDotNet \cite{ScriptHookVDotNet} and GTA V native. Information such as camera poses, GPS locations, and in-game time value are accessed through these libraries.
For our camera setup, we attached a pair of parallel cameras to the front hood of our ego vehicle 'Blista', with a stereo baseline of 0.54m. We set the vertical field of view to 59$^\circ$ and the horizontal field of view to 90$^\circ$. 
GTA V uses a distortion shader to make the game more realistic for players \cite{Courreges2015GTAVGraphicsStudy}. However, this shader poses potential issues for research, as the camera's distortion parameters are not publicly disclosed. Therefore, we remove the distortion shader to ensure undistorted stereo data output.

\subsection{TeFS: Temporal-controlled Frame Swap}

A potential solution for capturing stereo data in motion from GTA V is a simple frame swap method, where the left camera is captured in the first frame and the right camera in the second. However, this method does not account for temporal disparity between the two cameras, which can be up to 16.7ms for simulators running at 60 fps, and can easily double considering the rendering engine's one frame response time. While this issue is negligible for indoor or static scenes, it can cause significant position offsets (56-112cm for a car driving at 120km/h) when collecting stereo data from a moving vehicle. As a result, the captured image pairs are unsuitable for stereo odometry algorithms.

The primary challenge is the lack of per-frame temporal control over the camera-swapping process, similar to the issue noted in Hurl's PreSIL dataset \cite{hurl2019presil}. This limitation is due to the client-server architecture utilized by various GTA V-based data collection frameworks, including PreSIL, which prioritizes real-time analysis and scalability over low communication latency. To address this limitation, our proposed Temporal-controlled Frame Swap algorithm adopts a single-thread architecture that provides precise temporal control during the camera-swapping process,  as shown in the flow chart in \figref{TeFS_FC}.

\process

We introduce several important parameters and internal functions for our TeFS method. We use the parameter "cycleTick" to track in-game status changes during the stereo-collection cycle, which we denote as $T$ throughout the rest of the paper. The "onTick" function is automatically called by the game engine when the script is loaded, providing a fixed time step, or "uniTick," for tracking the global progress of the game engine, regardless of the frame rate. The "timeScale" parameter controls the passage of time in the game world, ranging from 0 to 1 and beyond, with 1 representing normal game speed and 0 representing a pseudo-pause, in which the entire scene remains stationary while the rendering engine still listens to function calls. The concept of pseudo-pause is crucial to our TeFS method since we cannot swap cameras while the game is natively paused. To avoid temporal disparity, pseudo-pause is our best replacement. However, pseudo-pause is not perfect, as many animations in GTA V are hard-coded, causing scenes to appear slowed down instead of completely halted. Therefore, to get the best results, we occasionally use the native pause function as well.

According to the process outlined in \figref{TeFS_FC}, we aim to collect one stereo image pair every 10 uniTicks once the data collection process commences. The ego vehicle drives normally for 7 uniTicks until $T+7$, at which point we set the timeScale parameter to 0 to achieve a pseudo-pause at the subsequent uniTick. At $T+8$, we pause the game and capture the RGB image for the left camera. Using the GTAVisionExport \cite{GTAVisionExport2018} repository, we access the DirectX depth buffer for the corresponding RGB frame to obtain depth information. We have modified GTAVisionExport to work seamlessly with our TeFS method. This ensures that we can store depth information with true scale units rather than normalized depth shader. Afterward, we resume the game and prepare for the camera swap.

After resuming the game, GTA V remains in the pseudo-pause state since we set the timeScale to 0 before the initial pause. This allows the rendering engine to execute our camera swap function while the scene remains relatively still. At $T+9$, we wait for the rendering engine to complete the camera-swapping process. When $T+10$ is reached, we can natively pause the game again and repeat the process for the right camera. At the end of $T+10$, we reset the timeScale to its normal speed, allowing the ego car to move forward and start the next data collection cycle.

\subsection{Temporal Information Conversion}
Capturing stereo data using the TeFS method involves a significant amount of temporal manipulation, so we cannot use real-world time to track the traversal timespan. Instead, we use GTA V's in-game time system to log each image pair's timestamp, as the in-game clock remains consistent even during significant slowdowns or native pauses throughout the data collection process. 
Based on our default stereo data capture setup, there is a 2.5 second in-game time difference between consecutive stereo image pairs, which translates to approximately 0.0825 seconds in real-world time. Therefore, we can further calculate our default camera frequency, CamFreq, to be approximately 12 fps using $\text{CamFreq} = \frac{1}{T_g \times \left(\frac{D_r}{86400}\right)}$, where $T_g$ is the in-game time passed between frames and $D_r$ is the real-world duration of one in-game day (default value is 48 minutes or 2880 seconds).

\subsection{Depth Ground Truth}

Depth information retrieved using GTAVisionExport \cite{GTAVisionExport2018} cannot be used directly because the stored information is in normalized device coordinates (NDC) format \cite{Racinsky2018}. It is a coordinate system used in computer graphics to represent positions on the viewport after the perspective projection and transformation have been applied. Followed by the work of Racinsky \cite{Racinsky2018} and PreSIL \cite{hurl2019presil}, we convert the given NDC value to each pixel's actual depth value by using $Depth = \frac{Map_{uv}}{NDC+{Map_{uv}} \times \frac{nc_z}{2 \times fc_z}}$, where $Depth$ represents the actual depth value of one pixel, and $NDC$ denotes the depth value in normalized device coordinates for the same pixel. $nc_z$ and $fc_z$ refer to the near and far clip distances of the camera, respectively. $Map_{uv}$ corresponds to the shape of the generated images and is a 2D array of values, with each element indicating the distance to a point on the near-clipping plane for a specific pixel (u,v) in the image. Furthermore, we also vectorized PreSIL's sequential generation approach, significantly enhancing the depth conversion process. 

\section{Experiments and  Evaluation}
Our experiments are conducted for two primary reasons. First, to demonstrate that the stereo data collected using our TeFS method is of similar quality to traditional dual-viewport methods, allowing state-of-the-art vSLAM algorithms to accurately estimate stereo trajectories. Second, to illustrate the value of the challenging-case comparison groups in our GTAV-TeFS dataset by pushing the state-of-the-art vSLAM models to their limits, effectively highlighting their unique strengths and limitations. Across the experiment section, we mainly use three metrics to evaluate the performance of various algorithms: Absolute Pose Error (APE, in both meters and percentages) and Relative Pose Error (RPE). APE(m) measures the global deviation from the ground truth for each pose in the trajectory, while APE(\%) helps eliminate the effect of trajectory length. RPE measures local motion accuracy, also known as drift. All metrics favor lower values.

\subsection{TeFS Validation with CARLA Implementation}
To validate the quality of our TeFS approach in stereo data collection and its universal applicability across various engines and games, we use the state-of-the-art open-source simulator, CARLA \cite{CARLA}, as our data comparison platform. Mirroring the steps taken with GTA V, we implemented the TeFS method in CARLA and established a temporal disparity of 0.5ms between the left and right camera, which is the lowest temporal disparity CARLA can support. Although it's not the same as the 0.2-0.3ms temporal disparity found in GTA V, they remain comparable to each other. Afterwards, we proceeded to collect two sets of stereo data: one using TeFS, and the other using the native stereo API. For each method we captured three subsets in separate maps, each consisting of 3600 images and yielding roughly 1km in trajectory length. 
With data ready, we performed stereo odometry evaluation on both datasets using ORB-SLAM3\cite{ORBSLAM3}. Both TeFS and native datasets illustrate excellent APE scores in all three predicted trajectory results(see \tabref{carlaValidate}). The maximum difference between two datasets' evaluation results is only 0.2\%, confirming that stereo data collected via TeFS are indeed on par with those derived from native stereo API.

\carlaValidate

\subsection{Challenging Case Evaluation for vSLAM Models}
Current state-of-the-art vSLAM models achieve great performance on established datasets such as KITTI\cite{kitti}. However, challenging cases such as hazardous environments, high noise, and low-light conditions still persist as obstacles for current vSLAM models. In the GTAV-TeFS dataset, we have prepared comparison data groups for three distinct road segments under various weather conditions, including one urban scenario and two off-road scenarios. Sample images can be found in  \figref{cgroup}. Each route has three corresponding weather combinations: 'Extra Sunny', 'Cloudy with Rain', and 'Night with Thunderstorm'. 'Extra Sunny' represents the basic condition,' Cloudy with Rain' serves as the moderate noise condition, and 'Night with Thunderstorm 'stands as the most challenging condition, featuring not only low contrast and high noise but also includes overexposure caused by lightning. All of the road segments incorporate loop closures, aiming to test the loop detection capabilities of each visual SLAM model. For models equipped with loop detection features, it should theoretically further enhance their evaluation results.

\aaaview

\subsubsection{Stereo Odometry Evaluation On Feature-based vSLAM Models}

\offroad

 We first undertook an evaluation of the stereo odometry capabilities of three feature-based vSLAM models, including ORB-SLAM3 \cite{ORBSLAM3}, VINS Fusion \cite{visualSlam}, and OV\begin{math}^{2}\end{math}SLAM\cite{qin2019b}. As presented in  \tabref{compare}, all three models show excellent performance in all road segments under the base 'Extra Sunny' weather. Notably, ORB-SLAM3 demonstrated superior robustness and accuracy in the longest road segment, Offroad 02, demonstrating 48\% lower APE than that of the other two. Moving to the rainy scenario, ORB-SLAM3 consistently outperformed the other visual SLAM models, with the mean APE score sitting comfortably below 0.35\%  maintaining its dominant edge across all road segments. This was largely attributed to its advanced loop merging capability; while VINS Fusion and OV2 both detected loop closure, the final merged map did not match the quality produced by ORB-SLAM3.

When subjected to the most challenging scenario, ‘Night with Thunderstorm’, all models suffered severe degradation in performance. ORB-SLAM3 lost a significant portion of the maps, and the results of the VINS FUSION across all three road segments were unusable. Remarkably, OV\begin{math}^{2}\end{math}SLAM was the only model that generated complete results for evaluation under these conditions, despite the high APE scores. Based on these evaluation results, we conclude that ORB-SLAM3 provides the most reliable performance and exceptional loop merging capability under normal and moderately noisy environments. Meanwhile, OV\begin{math}^{2}\end{math}SLAM demonstrated remarkable robustness, consistently delivering comparable maps under hazardous situations.

\subsubsection{Stereo and RGB-D Odometry Evaluation on Learning-based vSLAM Model}

In addition to feature-based vSLAM models, we also evaluated one learning-based model, DROID-SLAM \cite{teed2021droid}. Despite its claimed stereo support, the research community has reported multiple scaling issues with its stereo trajectory estimation results. Our evaluation, as presented in  \tabref{droid}, corroborates these findings. When only aligning the estimated trajectory with the origin without any scaling, the DROID-SLAM's stereo raw APE scores are excessively large, as if it is tested on monocular mode. Based on our observations over experiments, we need to match the scale of our prediction with ground truth. We first calculate scalar by dividing ground truth trajectory length with estimated length; then we multiply our estimations by the scalar. Upon scaling with ground truth, all trajectories produced by DROID-SLAM maintain the overall shape and demonstrate great accuracy. It is somewhat understandable that DROID-SLAM might encounter stereo scaling issues, considering the fact that the model was primarily trained on RGB-D data rather than stereo data.
\droid

Fortunately, our dataset also includes dense depth maps. Hence, to further substantiate our findings and assess the capabilities of DROID-SLAM, we incorporated RGB-D evaluation results alongside the stereo results. Unsurprisingly, both scaled stereo and RGB-D trajectory estimation under sunny weather conditions were readily handled by DROID-SLAM, with mean APE scores well below 0.3\% for both road segments. What surprised us was its performance in rainy and thunderstorm scenarios. Like other models, DROID-SLAM's stereo results suffered a performance downgrade under high-noise and overexposure environments. However, it still manages to produce much more consistent results than any feature-based vSLAM models we evaluated. Especially in the most challenging scenarios. To be specific, in both City 04 and Offroad 01's Night with Thunderstorm scene, previously only OV\begin{math}^{2}\end{math}SLAM was able to provide complete results with mean APE at 7.11\% and 7.57\%, respectively. While scaled stereo of DROID-SLAM provides mean APE at 0.84\% and 1.50\%, respectively.   Furthermore, its RGB-D results are even more outstanding, providing accurate trajectory estimation in all rainy and thunderstorm scenarios, with mean APE well below 0.64\%. However, DROID-SLAM has its own drawbacks as well, its long evaluation time and hardware requirement(11GB VRAM minimum \cite{teed2021droid}) are considerably demanding, causing some challenging and lengthy segments such as Offroad 02 in GTAV-TeFS dataset unable to finish evaluation before our RTX 3090 GPU (24GB VRAM) shut down the process. 





\section{TeFS Visual Comparisons}

 We present visual comparisons to showcase actual differences among the resulting outputs of the TeFS, native stereo, and the simple swap method through \figref{cgcompare}. The top images compare stereo data collected using TeFS and native API in CARLA. By visual comparison, the two appear identical due to the TeFS's minimal 0.5ms temporal offset, which translates to a 2mm spatial difference at 15km/h. This offset only causes a sub-pixel difference in a 1080p image. The bottom images contrast the stereo data collected via TeFS and the simple frame swap method in GTA V. The latter would have a temporal offset of 16.7ms,  which is 80 times larger than the 0.2ms temporal offset in TeFS. This significant discrepancy is causing the right output to lose important information and make depth-related tasks unfeasible.
\cgcompare

\section{Conclusion}
This paper introduces Temporal-controlled Frame Swap, a novel method designed to overcome the single viewport limitation in commercial video games like GTA V. By enabling the dynamic collection of high-quality stereo driving data, we creates the GTAV-TeFS dataset, which proves to be a valuable resource for evaluating and enhancing stereo vSLAM algorithms. The effectiveness and generalizability of the TeFS method, as well as the quality of the collected stereo data, have been scientifically analyzed and confirmed. With its precise control over in-game time, TeFS could potentially aid in the development of future surrounding-view data acquisition tools. This expansion of data collection capabilities presents exciting avenues for the advancement of future autonomy models.

\section{Acknowledgement}
Research was sponsored by the DEVCOM Analysis Center and was accomplished under Cooperative Agreement Number W911NF-22-2-0001. The views and conclusions contained in this document are those of the authors and should not be interpreted as representing the official policies, either expressed or implied, of the Army Research Office or the U.S. Government. The U.S. Government is authorized to reproduce and distribute reprints for Government purposes notwithstanding any copyright notation herein.


%
%

\bibliography{ref}

\begin{thebibliography}{22}
\providecommand{\natexlab}[1]{#1}
\providecommand{\url}[1]{\texttt{#1}}
\expandafter\ifx\csname urlstyle\endcsname\relax
  \providecommand{\doi}[1]{doi: #1}\else
  \providecommand{\doi}{doi: \begingroup \urlstyle{rm}\Url}\fi

\bibitem[Blade(2015)]{ScriptHookV}
Alexander Blade.
\newblock Scripthookv.
\newblock \url{http://www.dev-c.com/gtav/scripthookv}, 2015.

\bibitem[Campos et~al.(2021)Campos, Elvira, Gomez, Montiel, and
  Tardos]{ORBSLAM3}
Carlos Campos, Richard Elvira, Juan~J. Gomez, Jose M.~M. Montiel, and Juan~D.
  Tardos.
\newblock {ORB-SLAM3}: An accurate open-source library for visual,
  visual-inertial and multi-map {SLAM}.
\newblock \emph{IEEE Transactions on Robotics}, 37\penalty0 (6):\penalty0
  1874--1890, 2021.

\bibitem[Courrèges(2015)]{Courreges2015GTAVGraphicsStudy}
Adrian Courrèges.
\newblock Gta v graphics study.
\newblock
  \url{https://www.adriancourreges.com/blog/2015/11/02/gta-v-graphics-study/},
  2015.

\bibitem[Crosire(2015)]{ScriptHookVDotNet}
Crosire.
\newblock Scripthookvdotnet.
\newblock \url{https://github.com/crosire/scripthookvdotnet}, 2015.

\bibitem[Dosovitskiy et~al.(2017)Dosovitskiy, Ros, Codevilla, Lopez, and
  Koltun]{CARLA}
Alexey Dosovitskiy, German Ros, Felipe Codevilla, Antonio Lopez, and Vladlen
  Koltun.
\newblock Carla: An open urban driving simulator.
\newblock In \emph{Conference on robot learning}, pages 1--16. PMLR, 2017.

\bibitem[Geiger et~al.(2013)Geiger, Lenz, Stiller, and Urtasun]{kitti}
Andreas Geiger, Philip Lenz, Christoph Stiller, and Raquel Urtasun.
\newblock Vision meets robotics: The kitti dataset.
\newblock In \emph{International Journal of Robotics Research (IJRR)},
  volume~32, pages 1231--1237. Sage Publications Sage UK: London, England,
  2013.

\bibitem[Huang et~al.(2018)Huang, Matzen, Kopf, Ahuja, and Huang]{DeepMVS}
Po-Han Huang, Kevin Matzen, Johannes Kopf, Narendra Ahuja, and Jia-Bin Huang.
\newblock Deepmvs: Learning multi-view stereopsis.
\newblock In \emph{IEEE Conference on Computer Vision and Pattern Recognition
  (CVPR)}, 2018.

\bibitem[Hurl et~al.(2019)Hurl, Czarnecki, and Waslander]{hurl2019presil}
Braden Hurl, Krzysztof Czarnecki, and Steven Waslander.
\newblock Precise synthetic image and lidar (presil) dataset for autonomous
  vehicle perception.
\newblock In \emph{2019 IEEE Intelligent Vehicles Symposium (IV)}, pages
  2444--2451. IEEE, 2019.
\newblock \doi{10.1109/IVS.2019.8813809}.

\bibitem[Newcombe et~al.(2011)Newcombe, Lovegrove, and Davison]{visualSlam}
R.~A. Newcombe, S.~J. Lovegrove, and A.~J. Davison.
\newblock Dtam: Dense tracking and mapping in real-time.
\newblock In \emph{IEEE International Conference on Computer Vision (ICCV)},
  2011.
\newblock URL \url{https://ieeexplore.ieee.org/document/6126513}.

\bibitem[North(2013)]{gtav2013}
Rockstar North.
\newblock Grand theft auto v, 2013.
\newblock URL \url{https://www.rockstargames.com/V/}.

\bibitem[Qin et~al.(2019)Qin, Cao, Pan, and Shen]{qin2019b}
Tong Qin, Shaozu Cao, Jie Pan, and Shaojie Shen.
\newblock A general optimization-based framework for global pose estimation
  with multiple sensors, 2019.

\bibitem[Račinský(2018)]{Racinsky2018}
Matěj Račinský.
\newblock 3d map estimation from a single rgb image.
\newblock Master's thesis, Czech Technical University in Prague, May 2018.

\bibitem[Richter et~al.(2016)Richter, Vineet, Roth, and
  Koltun]{Richter2016PlayingForData}
Stephan~R. Richter, Vibhav Vineet, Stefan Roth, and Vladlen Koltun.
\newblock Playing for data: Ground truth from computer games.
\newblock In \emph{European Conference on Computer Vision (ECCV)}, pages
  102--118. Springer, 2016.

\bibitem[Richter et~al.(2017)Richter, Hayder, and
  Koltun]{Richter_2017playforbenchmark}
Stephan~R. Richter, Zeeshan Hayder, and Vladlen Koltun.
\newblock Playing for benchmarks.
\newblock In \emph{{IEEE} International Conference on Computer Vision, {ICCV}
  2017, Venice, Italy, October 22-29, 2017}, pages 2232--2241, 2017.
\newblock \doi{10.1109/ICCV.2017.243}.
\newblock URL \url{https://doi.org/10.1109/ICCV.2017.243}.

\bibitem[Ros et~al.(2016)Ros, Sellart, Materzynska, Vazquez, and
  Lopez]{synthia}
German Ros, Laura Sellart, Joanna Materzynska, David Vazquez, and Antonio~M
  Lopez.
\newblock The synthia dataset: A large collection of synthetic images for
  semantic segmentation of urban scenes.
\newblock In \emph{Proceedings of the IEEE conference on computer vision and
  pattern recognition}, pages 3234--3243, 2016.

\bibitem[ScreenRant(2019)]{gtav265}
ScreenRant.
\newblock Gta 5: How much it cost to make rockstar's open world game, 2019.
\newblock URL
  \url{https://screenrant.com/grand-theft-auto-5-how-much-cost-make/}.

\bibitem[Shah et~al.(2018)Shah, Dey, Lovett, and Kapoor]{airsim}
Shital Shah, Debadeepta Dey, Chris Lovett, and Ashish Kapoor.
\newblock Airsim: High-fidelity visual and physical simulation for autonomous
  vehicles.
\newblock In \emph{Field and Service Robotics}, pages 621--635. Springer, 2018.

\bibitem[Stewart and Waslander(2019)]{stewart2019ursa}
Andrew Stewart and Steven~L Waslander.
\newblock The ursa dataset: A large scale dataset for unmanned surface vehicle
  perception in coastal environments.
\newblock In \emph{2019 IEEE International Conference on Robotics and
  Automation (ICRA)}, pages 4603--4609. IEEE, 2019.
\newblock \doi{10.1109/ICRA.2019.8793719}.

\bibitem[Teed and Deng(2021)]{teed2021droid}
Zachary Teed and Jia Deng.
\newblock {DROID-SLAM: Deep Visual SLAM for Monocular, Stereo, and RGB-D
  Cameras}.
\newblock \emph{Advances in neural information processing systems}, 2021.

\bibitem[{UM \& Ford Center for Autonomous Vehicles
  (FCAV)}(2018)]{GTAVisionExport2018}
{UM \& Ford Center for Autonomous Vehicles (FCAV)}.
\newblock {GTAVisionExport}.
\newblock \url{https://github.com/umautobots/GTAVisionExport}, 2018.
\newblock GitHub repository.

\bibitem[Wang and Shen(2020)]{sfm}
Kaixuan Wang and Shaojie Shen.
\newblock Flow-motion and depth network for monocular stereo and beyond.
\newblock \emph{IEEE Robotics and Automation Letters}, 5\penalty0 (2):\penalty0
  3307--3314, 2020.
\newblock \doi{10.1109/LRA.2020.2975750}.

\bibitem[Wang et~al.(2020)Wang, Zhu, Wang, Hu, Qiu, Wang, Hu, Kapoor, and
  Scherer]{tartanair2020iros}
Wenshan Wang, Delong Zhu, Xiangwei Wang, Yaoyu Hu, Yuheng Qiu, Chen Wang, Yafei
  Hu, Ashish Kapoor, and Sebastian Scherer.
\newblock Tartanair: A dataset to push the limits of visual slam.
\newblock In \emph{2020 IEEE/RSJ International Conference on Intelligent Robots
  and Systems (IROS)}, pages 4909--4916. IEEE, 2020.

\end{thebibliography}
\end{document}